\documentclass[11pt,a4paper]{article}
\usepackage[hyperref]{acl2019}
\usepackage{times}
\usepackage{latexsym}

\usepackage{url}
\usepackage{amsmath}
\usepackage{multirow}
\usepackage{array}
\usepackage{graphicx}
\usepackage[normalem]{ulem}
\usepackage{rotating}

\usepackage{pgfplots}
\pgfplotsset{compat=1.7}
\usepackage{pgfplotstable}

\usepackage[T1]{fontenc}

\usepackage{csquotes}
\newcommand{\dq}[1]{\enquote{#1}}

\aclfinalcopy 
\def\aclpaperid{204} 

\def\rot#1{{\begin{sideways}#1\end{sideways}}}

\title{Efficiency Metrics for Data-Driven Models:\\ A Text Summarization Case Study}

\author{Erion \c{C}ano \\
  Institute of Formal and Applied \\
  Linguistics, Charles University, \\
  Prague, Czech Republic \\
  {\tt cano@ufal.mff.cuni.cz} \\\And
  Ond\v{r}ej Bojar \\
  Institute of Formal and Applied \\
  Linguistics, Charles University, \\
  Prague, Czech Republic \\
  {\tt bojar@ufal.mff.cuni.cz} \\}

\date{}

\begin{document}
\maketitle
\begin{abstract}
Using data-driven models for solving text summarization or similar tasks has become very common in the last years. Yet most of the studies report basic accuracy scores only, and nothing is known about the ability of the proposed models to improve when trained on more data. In this paper, we define and propose three data efficiency metrics: data score efficiency, data time deficiency and overall data efficiency. We also propose a simple scheme that uses those metrics and apply it for a more comprehensive evaluation of popular methods on text summarization and title generation tasks. For the latter task, we process and release a huge collection of 35 million abstract-title pairs from scientific articles. Our results reveal that among the tested models, the Transformer is the most efficient on both tasks. 
%
\end{abstract}

\section{Introduction}
%
Text summarization is the process of distilling the most noteworthy information in a document to produce an abridged version of it. This task is earning considerable interest, since shorter versions of long documents are easier to read and save us time.  
%
%
There are two basic ways to summarize texts. Extractive summarization selects the most relevant parts of the source document and combines them to generate the summary. In this case, the summary contains exact copies of words or phrases picked from the source. Abstractive summarization, on the other hand, paraphrases the information required for the summary instead of copying it verbatim. This is usually better, but also more complex and harder to achieve.
%

There has been a rapid progress in ATS (Abstractive Text Summarization) over the last years. The vanilla encoder-decoder with bidirectional LSTMs \cite{Hochreiter:1997:LSM:1246443.1246450} is now enhanced with advanced mechanisms like attention \cite{DBLP:journals/corr/BahdanauCB14} which has been widely embraced. It allows the model to focus on various parts of the input during the generation phase and was successfully used by \citet{D15-1044} to summarize news articles. Pointing (copying) is another mechanism that helps to alleviate the problem of unknown words \cite{gulcehre-etal-2016-pointing,gu-EtAl:2016:P16-1}. Moreover, coverage \cite{P16-1008} and intra-attention \cite{DBLP:journals/corr/PaulusXS17} were proposed and utilized to avoid word repetitions, producing more readable summaries. 
%
RL (Reinforcement Learning) concepts like policy gradient \cite{Rennie2017SelfCriticalST} were recently combined into the encoder-decoder architecture, alleviating other problems like train/test inconsistency and exposure bias \cite{DBLP:journals/corr/PaulusXS17,P18-1063}. 
%

All these developments helped to boost the ATS ROUGE \cite{Lin:2004} scores from about 30\,\% in \citet{D15-1044} to about 41\,\% in \citet{DBLP:journals/corr/PaulusXS17}. This is an increase of roughly 37\,\% in the last three years.  
%
Yet all the studies evaluate the methods using datasets of a fixed size. Doing so they tell us nothing about the expected performance\footnote{We use \dq{performance} solely for the output quality, not the time needed to train the model or obtain the output.} of the models when trained with more data. Moreover, training time is rarely reported. 
We believe that this evaluation practice of data-driven models is incomplete and data efficiency metrics should be computed and reported.

In this paper, we propose three data efficiency metrics, namely \emph{data score efficiency}, \emph{data time deficiency} and \emph{overall data efficiency}. The first two represent the output quality gain and the training time delay of the model per additional data samples. The third is the ratio between them and reflects the overall efficiency of the models w.r.t the training data. 
%
We also suggest a simple scheme that 
considers several values for each of the above metrics, together with the basic accuracy score, instead of reporting only the latter. The proposed scheme and the metrics can be used for a more detailed evaluation of supervised learning models.
%

Using them, we examine various recently proposed methods in two tasks: text summarization using the popular CNNDM (CNN/Daily Mail, \citealp{K16-1028}) dataset and title generation of scientific articles using OAGS, a novel dataset of abstract-title pairs that we processed and released.\footnote{\url{http://hdl.handle.net/11234/1-3043}} 
%
According to our results, the best-performing and fastest methods in the two datasets are those of \citet{DBLP:journals/corr/PaulusXS17} and \citet{P18-1063}. 
Regarding score and time efficiency, Transformer \cite{NIPS2017_7181} is distinctly superior. 
%
In the future, we will examine the Transformer model on more data with different parameter setups. Applying our evaluation scheme to related tasks such as MT (Machine Translation) could also be beneficial.
%

Overall, this work brings the following main contributions: (i) We define and propose three data efficiency metrics and a simple evaluation scheme that uses them for a more comprehensive evaluation of data-driven learning methods. (ii) We use the scheme and metrics to benchmark some of the most recently proposed ATS methods and discuss their training times, ROUGE, and data efficiency scores. (iii) Finally, a huge collection of about 35 million scientific paper abstracts and titles is prepared and released to the community. To our best knowledge, this is the largest data collection prepared for title generation experiments.
\section{Data Efficiency Metrics}  
\label{sec:metrics}
\subsection{Related Work}
\label{ssec:relwork}
Training data efficiency 
of the data-driven learning models is little considered in the literature. 
An early work is that of \citet{lawrence1998size} who investigate the generalization ability of neural networks with respect to the complexity of the approximation function, the size of the network and the degree of noise in the training data. In the case of latter factor, they vary the size of the training data and the levels of Gaussian noise added to those data concluding that ensemble techniques are more immune to the increased noise levels. Performance variations w.r.t the training data sizes are not considered, though. %

\citet{DBLP:journals/corr/Al-JarrahYMKT15} review the research literature focusing in the computational and energy efficiency of the data-driven methods. They particularly consider data-intensive application areas (e.g., big data computing) and how sustainable data models can help for a maximal learning accuracy with minimal computational cost and efficient processing of large volumes of data. %

\citet{DBLP:journals/corr/BoomLBSDD16} examine a character-level RNN (Recurrent Neural Network) used to predict the next character of a text given the previous input characters. They assess the evolution of the network performance (in terms of perplexity) in four train and prediction scenarios as a function of the training time and input training sequences. According to their results, the efficiency of the model is considerably influenced by the chosen scenario.  
%

A similar experiment is conducted by \citet{riou:hal-02022678} who explore reinforcement learning concepts on the task of neural language generation. They compare different implementations reporting not only performance scores, but also their evolution as a function of the cumulated learning cost and the training data size.  
%

The most relevant work we found is the one by \citet{icpram19} who propose an experimental protocol for comparing the data efficiency of a CNN (Convolution Neural Network) with that of HiGSFA (Hierarchical information-preserving Graph-based Slow Feature Analysis). They give an informal definition of data efficiency considering it as \emph{performance as a function of training set size}. Three character recognition challenges are defined and the two methods are trained on increasing amounts of data samples reporting the corresponding accuracy scores.
\subsection{Proposed Data Efficiency Metrics}
\label{ssec:metrics}
%
Despite the experimental results and insights they bring, the above studies are still task and method specific. Moreover, their computation schemes are not generic or transferable and no formalization of the data efficiency is given. In this section, we define three novel and useful data efficiency metrics.      
%

Suppose we train a data-driven method \textbf{M} on dataset \textbf{D} to solve task \textbf{T} and we test it based on performance score \textbf{S}. We also assume that the quality of the data samples in different intervals of \textbf{D} is homogeneous. In practice, this could be achieved by shuffling \textbf{D} before starting the experiments. For a certain training data size $d$, it takes $t$ seconds to train the model $m_d$ until convergence (i.e. until no further gains are observed with more training time) and the score obtained by testing it on a standard and independent test dataset of a fixed size is $s$. We expect that for a certain increase $\Delta d$ of training samples fed to {\bf M}, it will require an extra time $\Delta t$ to converge, and the resulting model $m_{d + \Delta d}$ will attain an extra $\Delta s$ score. We can thus define and compute \emph{data score efficiency} (score gain per additional data samples) $\Sigma$ of method \textbf{M} as:
\begin{equation}
\label{eq:Score}
\Sigma~=~\Delta s~/~\Delta d
\end{equation}
It is a measure of how smartly or effectively \textbf{M} interprets the extra data samples, or how well its performance score scales w.r.t the training data. 
%
%
Similarly, \emph{data time deficiency} (the inverse of \emph{data time efficiency}) $\Theta$ of \textbf{M} will be:
\begin{equation}
\label{eq:Time}
\Theta~=~\Delta t~/~\Delta d
\end{equation} 
This measures how slowly or lazily \textbf{M} interprets the additional samples.\footnote{Our \emph{data time efficiency} ($\Delta d$~/~$\Delta t$) should not be confused with the \emph{training throughput} as defined by \citet{popel:bojar:TrainingTipsfortheTransformerModel:2018} for machine translation which reflects the time required for one model update given the additional data. Our $\Delta t$ is the increase in the overall training time till convergence on the enlarged dataset in comparison with the original one.}
Given two train and test runs (original and enlarged datasets) characterized by the above measures (training data: $d$, $d + \Delta d$; training times: $t$, $t + \Delta t$; achieved scores: $s$, $s + \Delta s$), we define the \emph{overall data efficiency} $E$ as: 
\begin{equation}
\label{eq:Data}
E~=~\Sigma~/~\Theta~=~\Delta s~/~\Delta t
\end{equation}
It is a measure of how smartly and quickly the models of {\bf M} utilizes the data of {\bf D} on task {\bf T}.
%

In practice, using the absolute increments $\Delta s,~\Delta t,$ and $\Delta d$ may produce small values of $\Sigma$ which are hard to interpret and work with. Furthermore, $\Theta$ and $E$ use training times which depend on the computing conditions (e.g., hardware setups). As a result, they are hardly reproducible across different computing environments.
%
%
To overcome these limitations, we can instead use the relative increments $\Delta s/s,~\Delta t/t$ and $\Delta d/d$, computing the corresponding \emph{relative data efficiency metrics} as: 
%
\begin{align}
\label{eq:score}
\sigma &= \frac{\Delta s~/~s}{\Delta d~/~d} 
\\ 
%
\label{eq:time}
\theta &= \frac{\Delta t~/~t}{\Delta d~/~d} 
\\
%
\label{eq:data}
\epsilon = &\frac{\sigma}{\theta} = \frac{\Delta s~/~s}{\Delta t~/~t} 
\end{align}
%
%
These relative metrics and their values are practically easier to interpret and work with. Furthermore, they are transferable or reproducible in different computing setups which is important for cross-interpretation of the experimental results. We can express $\sigma$ and $\theta$ values in percent and $\epsilon$ values as their ratio.
\subsection{Assorted Remarks}
\label{ssec:remarks}
The metrics presented above can be used to evaluate different data-driven methods or compare several parameter configurations of the same basic method (algorithm, neural network, etc.) and help us find the optimal one. In this sense, they are generic and task-independent.
%
However, it is important to note that they do not represent \dq{universal} or global attributes of method \textbf{M}. 
They are instead linear approximations that can give us local characterizations of \textbf{M} in certain intervals of \textbf{D}. In other words, high $\Sigma$ (or $\sigma$) values of \textbf{M} in some intervals of \textbf{D} do not necessarily assure a decent generalization of \textbf{M}. 
%

It is also important not to confuse the data efficiency with performance or quality. In our daily intuition, we often tend to consider highly efficient machines, techniques or methods as well-performing ones. Instead, according to the above definitions, a model can perform poorly but still be highly efficient w.r.t the training data. This happens if its performance scores on increasing training data cuts are all very low, but grow very quickly from one assessment to the next. A model can also yield high scores which grow very slowly on increasing data sizes (thus relatively small $\Sigma$ and $\sigma$ values). In this case it is a well-performing (maybe even the best) model on those data, but not a data efficient one. 
%

From the data efficiency viewpoint, the best models would obviously be those of higher \emph{data score efficiency} and lower \emph{data time deficiency}, or higher \emph{overall data efficiency}. In practice, performance is generally the most desired characteristic. As a result, \emph{data score efficiency} values ($\Sigma$, $\sigma$ or both) should be more important and worthy to report in most of the cases. 
%
%
Since models are trained only once, $\theta$ and $\epsilon$ should be less relevant. Nevertheless, they might be useful from a technical or theoretical perspective. They can be used for comparing different methods, comparing different parameter configurations of a method, or for trying run time optimizations.   
%
%
\section{A Comprehensive Evaluation Scheme} 
\label{sec:evalscheme}
%
Since the sizes of the predictive models and the utilized datasets are consistently growing, it becomes more difficult and costly to use human expertise for the evaluation. The typical approach is to test automatically by means of standard datasets and scoring metrics which are popular. For example, in the case of text summarization task, it is very common to find evaluations of proposed methods using the full set of CNNDM only (Table~1 in \citeauthor{DBLP:journals/corr/PaulusXS17}, Table~3 in \citeauthor{P18-2027}, Table~1 in \citeauthor{P17-1099}, and more).  

We believe there are serious shortcomings in this evaluation practice.
Testing only one model of a method trained on a fixed-size data split does not reveal anything about its score trend when fed with more data. It thus becomes hard to discern the overall best method (out of a few that are compared) in a fair and objective way, especially if the achieved scores are similar. Moreover, training time is rarely reported and nothing is known about the time efficiency of the models. 
%

To overcome the above limitations, we propose a more detailed evaluation scheme that considers accuracy scores together with the data efficiency metrics defined in Section~\ref{ssec:metrics}. Again, suppose we have a dataset \textbf{D} of size $d$ with homogeneous training samples, a standard performance score \textbf{S} and two methods \textbf{A} and \textbf{B} that we want to compare. The typical practice trains two single models $a$ and $b$ from \textbf{A} and \textbf{B}  
on entire $d$ and reports accuracy scores $s^a$ and $s^b$ from the standard test set. 
%

Instead, we suggest to split $d$ in $n$ equal parts of size $d/n$ and form $n$ intervals $d_1,~d_2,~\ldots,~d_n$ of increasing sizes $d/n,~2d/n,~\ldots,~(n-1)d/n,~d$. This way we can train $2n$ models $a_1,~a_2, ~\ldots,~a_n$ and $b_1,~b_2, ~\ldots,~b_n$ on $d_1,~d_2,~\ldots,~d_n$ and compute their scores $s_1^a,~s_2^a, ~\ldots,~s_n^a$ and $s_1^b,~s_2^b, ~\ldots,~s_n^b$ from the same test set. From Equation~\ref{eq:score}, we also compute $\sigma_1^a,~\sigma_2^a,~\ldots,~\sigma_{n-1}^a$ using each two scores $s_i^a$ and $s_{i+1}^a$ of models $a_i$ and $a_{i+1}$, together with $\sigma_1^b,~\sigma_2^b,~\ldots,~\sigma_{n-1}^b$ from the \textbf{B} models.
%

We can now report up to $2n$ score values and $2(n-1)$ relative data score efficiency values. For conciseness, we can limit in $s_n^a$ and $s_n^b$ of the two biggest models. Also, given the local nature of the efficiency metrics, it make sense to report values from dispersed data intervals like the leftmost ($\sigma_1^a$ and $\sigma_1^b$), the middle ($\sigma_{n/2}^a$ and $\sigma_{n/2}^b$) and the rightmost ($\sigma_{n-1}^a$ and $\sigma_{n-1}^b$) $\sigma$. The rightmost values are probably more relevant for predicting the score trend on bigger training sizes. We can also compute and report the respective $\Sigma$ values or even the $\theta$ and $\epsilon$ values in a similar fashion using the other equations of Section~\ref{ssec:metrics}.
%
%
%

\begin{figure}[!t]
\centering
\includegraphics[width=0.97\linewidth]{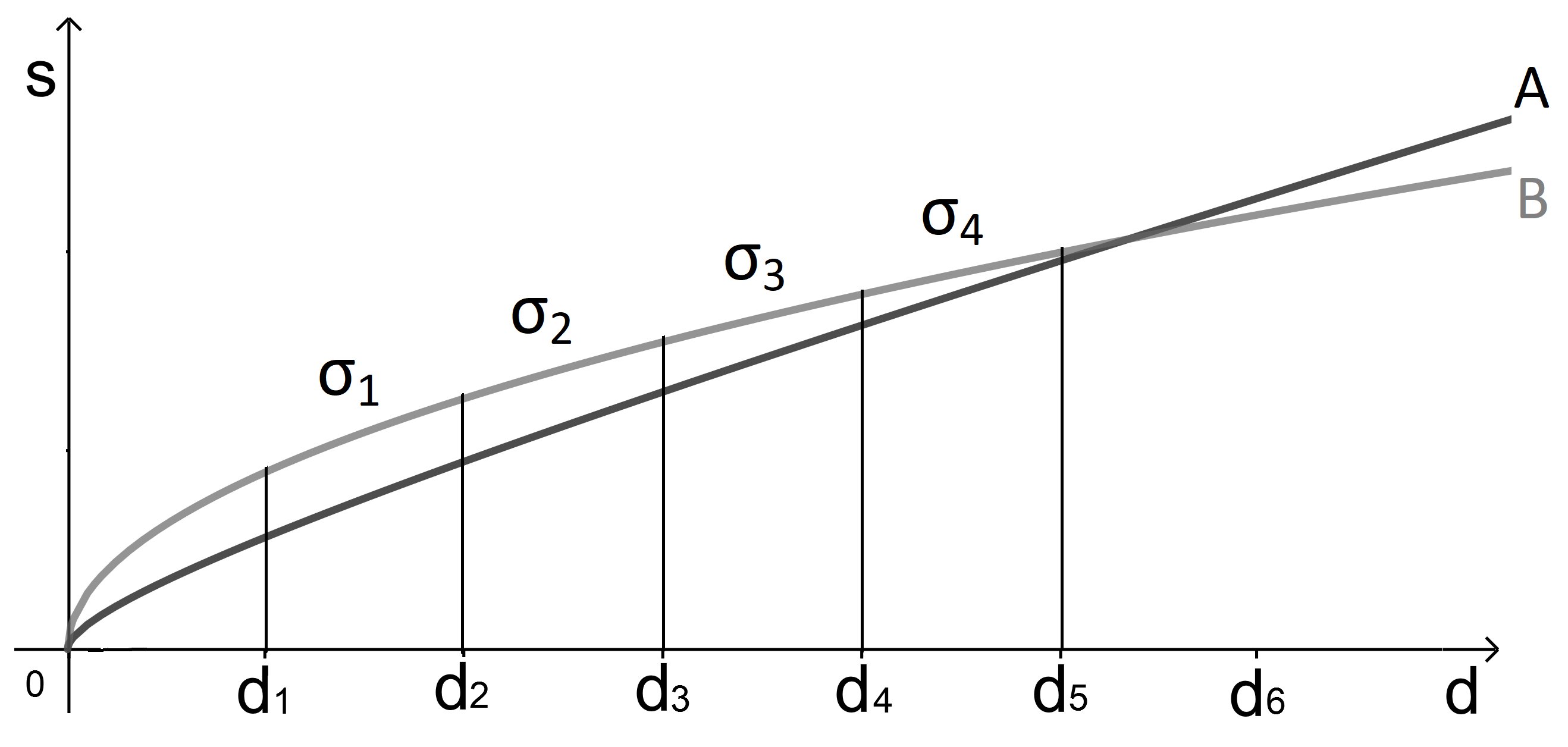}
\caption{Illustration of the schema application}
\label{fig:scalability}
\end{figure}
%

Getting back to \textbf{A} vs. \textbf{B}, we can first check $s_n^a$ and $s_n^b$. If one of them is distinctly higher than the other, comparing the $\sigma$ values may not be essential. The real worth comes when $s_n^a \approx s_n^b$, by contrasting the rightmost corresponding $\sigma$ values ($\sigma_{n-1}^a$ vs. $\sigma_{n-1}^b$). A significant difference of one against the other could suggest which of them will reach higher scores on a bigger training set.
%

To illustrate, we can see in Figure~\ref{fig:scalability} two hypothetical graphs that approximate the variations of $s^a$ and $s^b$ over \textbf{D}. We have $n = 5$, training size $d = d_5$ and very similar performance scores ($s_5^a \approx s_5^b$). Obviously, $s^b$ grows faster than $s^a$ till $d_2$, but then the situation is reversed, since $\sigma_3^a > \sigma_3^b$ and $\sigma_4^a > \sigma_4^b$. We can thus expect $s^a > s^b$ for $d > d_5$ which is what actually happens in this example ($s_6^a > s_6^b$).  
%

Using the traditional practice (computing $s_5^a$ and $s_5^b$ only) our verdict would be: \emph{\textbf{A} and \textbf{B} perform (almost) the same on \textbf{D}}. Instead, using the above scheme we can conclude that: \emph{\textbf{A} and \textbf{B} perform (almost) the same on \textbf{D}, but \textbf{A} will probably perform better than \textbf{B} if trained on more data}. The scheme can be used to evaluate data-driven methods with different scores, on different tasks. In Section~\ref{sec:experiments} we show the results we obtained by applying it 
to assess several advanced ATS methods.    
\section{Text Summarization Datasets} 
\label{sec:data}
%
The tendency towards data-driven methods based on neural networks has encouraged experiments with large text collections for various tasks. In the case of ATS, one of the first big datasets was the annotated English Gigaword \cite{Napoles:2012:AG:2391200.2391218,D15-1044} with over nine million news articles and headlines processed using CoreNLP of  \citet{manning-EtAl:2014:P14-5}. Each headline was paired with the first sentence of the corresponding article to create the training base for the experiments. 
%
DUC-2004 is another dataset\footnote{\url{https://duc.nist.gov/duc2004/}}, mostly used as an evaluation baseline, given its small size. It consists of 500 document-summary pairs curated by human experts.   
%
Newsroom is a recent and heterogeneous bundle of about 1.3 million news articles \cite{N18-1065}. 
%

CNNDM has become the most popular dataset for text summarization \cite{K16-1028}. It provides a large set of news articles and the corresponding multi-sentence summaries, unlike the three above that contain one-sentence summaries only. It is thus more suitable for training and testing summarization models of longer texts.
%

Title generation task, on the other hand, requires data samples of shorter texts and one-sentence titles. Collections of abstracts and titles from scientific articles are well suited for exploring it.     
%
KP20k is a collection of 20K records of scientific paper metadata (title, abstract and keywords) presented by \citet{P17-1054}. 
The metadata belong to articles of computer science from ACM Digital Library, ScienceDirect, and Web of Science.   
%
%
\begin{table}[!t]
	\centering
	\small
	\newcolumntype{C}[1]{>{\centering\arraybackslash} m{#1}}
	\begin{tabular}{|c|c|r|c|c|r|r|r|}
		\hline
		\strut\bf  & \bf Split & \bf Rec & \bf SrcL & \bf TgtL & \bf Voc & \bf Used \\ 
		\hline
		\multirow{5}{*}{\rot{CNNDM}} 
		& Train1 & 96K & 784 & 54 & 380K & 49K \\ 
		& Train2 & 192K & 780 & 57 & 555K & 49K \\ 
		& Train3 & 287K & 786 & 55 & 690K & 49K \\ 
		& Valid & 13K & 769 & 61 & -- & -- \\ 
		& Test & 11K & 787 & 58 & -- & -- \\ 
		\hline
		\multirow{5}{*}{\rot{OAGS}} 
		& Train1 & 500K & 183 & 9 & 1.2M & 98K \\ 
		& Train2 & 1M & 205 & 10 & 2.1M & 98K \\ 
		& Train3 & 1.5M & 211 & 11 & 2.8M & 98K \\ 
		& Valid & 10K & 231 & 13 & -- & -- \\ 
		& Test & 10K & 237 & 12 & -- & -- \\ 
		\hline
	\end{tabular}
	\caption{Statistics of used datasets. For each split, it shows the number of
	records (Rec), average length of source and target texts in tokens (SrcL,
	TgtL), total vocabulary size (Voc), and the number of most frequent words
	that were used (Used).}
	\label{table:datastats}
\end{table}
%

The demand for more and more data has motivated initiatives that mine research articles from academic networks. One of them is ArnetMiner, a system that extracts researcher profiles from the Web and integrates the data into a unified network \cite{tang2008arnetminer}. A byproduct of that work is the OAG (Open Academic Graph) collection \cite{Sinha:2015:OMA:2740908.2742839}.
%

To produce a big title generation dataset for our experiments, we started from OAG. First, \emph{abstract}, \emph{title}, and \emph{language} fields were extracted from each record where they were available. In many cases, abstract language did not match the \emph{language} field. We ignored the latter and used a language identifier to remove records that were not in English. Duplicates were dropped and the texts were lowercased. Finally, Stanford CoreNLP tokenizer was used to split title and abstract texts. 
%
The resulting dataset (OAGS, released with this paper) contains about 35 million abstract-title pairs and can be used for title generation experiments.
%

We had a quick look at the content of OAGS and observed that most of the papers are from medicine. There are also many papers about social sciences, psychology, economics or engineering disciplines. Given its huge size and the topical richness, the value of OAGS is twofold: (i) It can be used to supplement existing datasets on title generation tasks when more training data are needed. (ii) It can be used for creating byproducts of specific scientific disciplines or domains.     
%

%

\section{Text Summarization Evaluation}
\label{sec:experiments}
%
In this section, we apply the relative metrics of Section~\ref{ssec:metrics} and the evaluation scheme of Section~\ref{sec:evalscheme} to benchmark several advanced methods on text summarization of news articles and title generation of scientific papers. We first introduce the methods and their parameters, together with the experimental data. Later, we present and discuss the achieved accuracy and data efficiency scores. 
%

\subsection{Tested Summarization Methods} 
\label{ssec:absts}
%
The ability of recurrent neural networks to represent and process variable-length sequences has created a tradition of applying them on sequence-to-sequence tasks such as ATS or MT. In the case of ATS, the goal is to process the source text producing a target text that is shorter but still meaningful and easy to read. 
%
%

\citet{D15-1044} were probably the first to implement attention in a network dedicated to ATS.
Their model (\textsc{Abs} in the following) uses an encoder that learns a soft alignment (attention) between the source and the target sequences producing the context vector. In the decoding phase, it uses a beam-search decoder \cite{Dahlmeier:2012:BDG:2390948.2391013} with a window of 10 candidate words in each target position. There are 256 and 128 dimensions in the hidden layer and word embedding layer respectively. The authors reported state-of-the-art results in the DUC-2004 testing dataset.
%

\citet{P17-1099} proposed Pointer-Generator (\textsc{Pcov}), a model that implements an attention-based encoder for producing the context vector.
The decoder is extended with a pointing/copying mechanism \cite{gulcehre-etal-2016-pointing,gu-EtAl:2016:P16-1} that is used in each step to compute a generation probability $p_{gen}$ from the context vector, the decoder states, and the decoder output in that step. This generation probability is used as a switch to decide if the next word should be predicted or copied from the input. 
Another extension is the coverage mechanism (keeping track of decoder outputs) for avoiding word repetitions in the summary, a chronic problem of encoder-decoder summarizers \cite{P16-1008}. The method was implemented with word embeddings and hidden layer of sizes 128 and 256 respectively. 
%

\begin{table*}[!t]
\small 
\centering      
\setlength\tabcolsep{6pt}  
\begin{tabular}
{| l l | c c c c c | c c c c c |}
\hline
& \multicolumn{11}{c|}{\qquad \qquad \qquad \qquad \textbf{CNNDM} \qquad \qquad \qquad \qquad \qquad \qquad \textbf{OAGS}} \\ [0.12ex] 
\textbf{Authors} & \textbf{Model} & $\boldsymbol{P}$ & $\boldsymbol{R_1}$ & $\boldsymbol{R_2}$ & 
$\boldsymbol{R_L}$ & $\boldsymbol{T_t}$ & $\boldsymbol{P}$ & $\boldsymbol{R_1}$ & 
$\boldsymbol{R_2}$ & $\boldsymbol{R_L}$ & $\boldsymbol{T_t}$ \\ [0.12ex] 
\hline
\multirow{3}{*}{\citeauthor{D15-1044}}
& \textsc{Abs1} & 15M & 26.66 & 8.81 & 24.46 & 135032 & 22M & 24.75 & 10.05 & 21.84 & 48595 \\ [0.07ex]
& \textsc{Abs2} & 15M & 28.56 & 10.42 & 25.57 & 185549 & 22M & 26.6 & 11.5 & 23.33 & 61729 \\ [0.07ex]
& \textsc{Abs3} & 15M & 29.64 & 11.55 & 26.32 & 243549 & 22M & 27.86 & 12.15 & 24.48 & 73038 \\ [0.77ex]
\multirow{3}{*}{\citeauthor{P17-1099}}
& \textsc{Pcov1} & 14M & 36.97 & 15.19 & 33.84 & 113110 & 21M & 34.4 & 17.67 & 27.55 & {\bf 30551} \\ [0.07ex]
& \textsc{Pcov2} & 14M & 38.56 & 16.03 & 35.09 & 138175 & 21M & 35.18 & 18.06 & 28.83 & 42723 \\ [0.07ex]
& \textsc{Pcov3} & 14M & 39.41 & 16.77 & 36.31 & 163744 & 21M & 35.86 & 18.51 & 29.42 & 56538 \\ [0.77ex]
\multirow{3}{*}{\citeauthor{DBLP:journals/corr/abs-1812-02303}}
& \textsc{Nats1} & 15M & 36.92 & 14.56 & 32.88 & 98791 & -- & -- & -- & -- & -- \\ [0.07ex]
& \textsc{Nats2} & 15M & 38.25 & 15.89 & 34.02 & 179689 & -- & -- & -- & -- & -- \\ [0.07ex]
& \textsc{Nats3} & 15M & 39.11 & 17.2 & 35.66 & 261794 & -- & -- & -- & -- & -- \\ [0.77ex]
\multirow{3}{*}{\citeauthor{P18-2027}}
& \textsc{GlobEn1} & 68M & 36.53 & 14.9 & 34.11 & 658924 & -- & -- & -- & -- & -- \\ [0.07ex]
& \textsc{GlobEn2} & 68M & 37.82 & 16.13 & 35.46 & 785622 & -- & -- & -- & -- & -- \\ [0.07ex]
& \textsc{GlobEn3} & 68M & 38.67 & 16.94 & 36.25 & 875817 & -- & -- & -- & -- & -- \\ [0.77ex]
\multirow{3}{*}{\citeauthor{NIPS2017_7181}}
& \textsc{Trans1} & 81M & 32.38 & 10.47 & 29.43 & 518924 & 129M & 30.29 & 13.1 & 24.34 & 251802 \\ [0.07ex]
& \textsc{Trans2} & 81M & 36.76 & 14.54 & 33.82 & 579149 & 129M & 34.17 & 17.49 & 28.46 & 269665 \\ [0.07ex]
& \textsc{Trans3} & 81M & 38.24 & 16.33 & 35.28 & 611359 & 129M & 37.06 & {\bf 19.44} & 30.51 & 278602 \\ [0.77ex]
\multirow{3}{*}{\hyperlink{P18-1063}{Chen et al.}} 
& \textsc{FastRl1} & -- & 36.95 & 14.89 & 34.69 & {\bf 19601} & -- & -- & -- & -- & -- \\ [0.07ex]
& \textsc{FastRl2} & -- & 39.18 & 16.17 & 36.15 & 30485 & -- & -- & -- & -- & -- \\ [0.07ex]
& \textsc{FastRl3} & -- & 40.02 & {\bf 17.52} & 37.24 & 52775 & -- & -- & -- & -- & -- \\ [0.77ex]
\multirow{3}{*}{\citeauthor{DBLP:journals/corr/PaulusXS17}}
& \textsc{Pgrl1} & -- & 38.16 & 14.17 & 36.24 & 68942 & -- & 35.52 & 16.81 & 28.65 & 43726 \\ [0.07ex]
& \textsc{Pgrl2} & -- & 39.88 & 15.31 & 37.89 & 81529 & -- & 36.9 & 18.44 & 30.22 & 55324 \\ [0.07ex]
& \textsc{Pgrl3} & -- & {\bf 40.83} & 15.68 & {\bf 38.73} & 107179 & -- & {\bf 38.05} & 19.23 & {\bf 31.16} & 74983 \\ [0.77ex]
\hline 
\end{tabular}
\caption{Parameters, ROUGE $F_1$ scores and training times for each method on the splits of the two datasets} 
\label{table:scores}
\end{table*}
%

\begin{figure*}[!t]
\centering
\begin{minipage}{0.49\textwidth}
\pgfplotstableread[col sep=&, header=true]{
description & ABS & PCOV & NATS & GLOBEN & TRANS & FASTRL & PGRL
1 & 26.66 & 36.97 & 36.92 & 36.53 & 32.38 & 36.95 & 38.16
2 & 28.56 & 38.56 & 38.25 & 37.82 & 36.76 & 39.18 & 39.88
3 & 29.64 & 39.41 & 39.11 & 38.67 & 38.24 & 40.02 & 40.83
}\datatableentry
\begin{tikzpicture}[scale=0.94] 
\begin{axis}[
  title={},
  xtick=data,
  xticklabels ={1, 2, 3},
  x tick label style={font=\footnotesize},
  ytick={5, 10, 15, 20, 25, 30, 35, 40, 45},
  ylabel={R-1 score},
  y tick label style={font=\footnotesize},
  legend style={font=\tiny,legend pos=outer north east},
  ymajorgrids=true,
  grid style=dashed,
  legend to name=savedlegend,  
  legend columns=7, 
]
\addlegendentry{ABS};
\addplot [thick, color=gray, mark=square] table [y=ABS, x expr=\coordindex] {\datatableentry};
\addlegendentry{PCOV};
\addplot [thick, color=blue, mark=triangle] table [y=PCOV, x expr=\coordindex] {\datatableentry};
\addlegendentry{NATS};
\addplot [color=black, mark=asterisk] table [y=NATS, x expr=\coordindex] {\datatableentry};
\addlegendentry{GLOBEN};
\addplot [thick, color=yellow, mark=oplus] table [y=GLOBEN, x expr=\coordindex] {\datatableentry};
\addlegendentry{TRANS};
\addplot [thick, color=green, mark=diamond] table [y=TRANS, x expr=\coordindex] {\datatableentry};
\addlegendentry{FASTRL};
\addplot [thick, color=red, mark=halfcircle] table [y=FASTRL, x expr=\coordindex] {\datatableentry};
\addlegendentry{PGRL};
\addplot [thick, color=orange, mark=pentagon] table [y=PGRL, x expr=\coordindex] {\datatableentry};
\end{axis}
\end{tikzpicture} 
\end{minipage}
%
\begin{minipage}{0.49\textwidth}
\pgfplotstableread[col sep=&, header=true]{
description & ABS & PCOV & TRANS & PGRL
1 & 24.75 & 34.4 & 30.29 & 35.52 
2 & 26.6 & 35.18 & 34.17 & 36.9
3 & 27.86 & 35.86 & 37.06 & 38.05
}\datatableentry
\begin{tikzpicture}[scale=0.94] 
\begin{axis}[
  title={},
  xtick=data,
  xticklabels ={1, 2, 3},
  x tick label style={font=\footnotesize},
  ytick={5, 10, 15, 20, 25, 30, 35, 40, 45},
  y tick label style={font=\footnotesize},
  ymajorgrids=true,
  grid style=dashed,
  legend style={font=\footnotesize,legend pos=outer north east},
]
\addlegendentry{ABS};
\addplot [thick, color=gray, mark=square] table [y=ABS, x expr=\coordindex] {\datatableentry};
\addlegendentry{PCOV};
\addplot [thick, color=blue, mark=triangle] table [y=PCOV, x expr=\coordindex] {\datatableentry};
\addlegendentry{TRANS};
\addplot [thick, color=green, mark=diamond] table [y=TRANS, x expr=\coordindex] {\datatableentry};
\addlegendentry{PGRL};
\addplot [thick, color=orange, mark=pentagon] table [y=PGRL, x expr=\coordindex] {\datatableentry};
\legend{}; 
\end{axis}
\end{tikzpicture} 
\end{minipage}
\ref{savedlegend}  
\caption{$\boldsymbol{R_1}$ score trends of the three models of each method on CNNDM (left) and OAGS (right)}
\label{fig:r1trends}
\end{figure*}
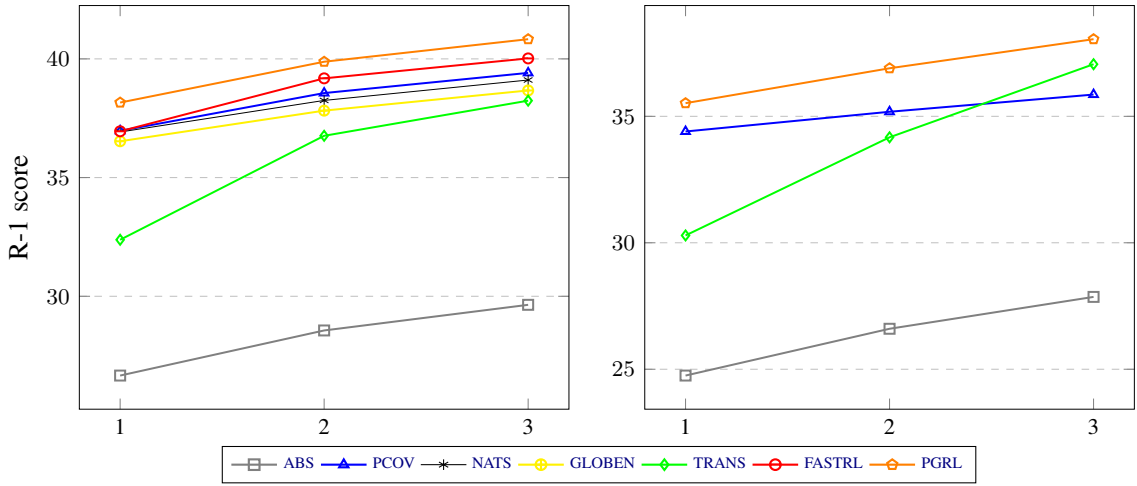
%

%

\citet{P18-2027} tried a partial use of convolutions in their model (\textsc{GlobEn}) to avoid word repetitions and semantic irrelevance in the summaries.
They couple the encoder with a convolutional gated unit which performs global encoding of the source context and uses it to filter certain n-gram features and refine the output of the encoder in each time step. \textsc{GlobEn} is a very big network (about 68M parameters on CNNDM) with three layers in the encoder and other three in the decoder, each of 512 dimensions. 
%

A taxonomy of the above (and more) sequence-to-sequence methods and added mechanisms can be found in \citet{DBLP:journals/corr/abs-1812-02303}. Authors present a detailed review of problems and proposed solutions based on network structures, training strategies, and generation algorithms. Furthermore, they develop and release a library (\textsc{Nats}) that implements combinations of mechanisms like attention, pointing, and coverage, analyzing their effects in text summarization quality.
\textsc{Nats} was implemented with the same network parameters as \textsc{Pcov}. Intra-decoder attention and weight sharing of embeddings were added in the decoder. 
%

The introduction of the Transformer (\textsc{Trans}) architecture that removes all recurrent or convolutional structures reduced computation cost and training time \cite{NIPS2017_7181}. Totally based on attention mechanism and primarily designed for MT, Transformer can also work for text summarization, since all it needs to do is to learn the alignments between the input (source) texts and the output (target) summaries. Positional encoding is added to word embeddings to preserve the order of the input and output sequences. 
\textsc{Trans} is the biggest model we tried, with four layers in both encoder and decoder, 512 dimensions in each layer, including the embedding layers, 200K training steps and 8000 warm-up steps.

Two observed problems in the encoder-decoder framework are the \emph{exposure bias} and \emph{train/test inconsistency} \cite{DBLP:journals/corr/abs-1805-09461}. 
%
To overcome them, RL ideas have been recently applied. \citet{DBLP:journals/corr/PaulusXS17} use intra-attention to focus on different parts of the encoded sequence.
This way it is less likely for their model (\textsc{Pgrl}) to attend to the same parts of input in different decoding steps, and thus fewer word repetitions should appear in the summaries. 
%
To optimize for ROUGE or similar discrete evaluation metrics, they implement self-critical policy gradient training with reward function, a RL mechanism introduced by \citet{Rennie2017SelfCriticalST}. 
\textsc{Pgrl} was used with encoder and decoder of 256 dimensions and word embeddings of 128 dimensions. 

Aiming for speed, \citet{P18-1063} developed an extractive-abstractive text summarizer (\textsc{FastRl}) with policy-based reinforcement.
It first uses an extractor agent to pick the most salient sentences or phrases, instead of encoding the entire input sequence which can be long. It then uses an encoder-decoder abstractor to rewrite (compress) the sentences in parallel. Actor-Critic policy gradient with reward function \cite{bahdanau+al-2016-actorcritic} joins together the extractor and abstractor networks. 
Same as most models above, \textsc{FastRl} uses 256 and 128 dimensions for the recurrent layer and the word embeddings. 
 
In every experiment, no pretraining of word embeddings was performed. They were learned during the training of each model. Adam optimizer \cite{Adam} was used with $\alpha = 0.001,~\beta_1 = 0.9,~\beta_2 = 0.999$ and $\epsilon = 10^{-8}$. 
We chose mini-batches of size 16 in most of the cases (8 for \textsc{GlobEn} and \textsc{Trans} to avoid memory errors). All experiments were conducted on two NVIDIA GTX 1080Ti GPUs. 
%
%
\subsection{Used Data}
\label{ssec:usedData}
To cope with limited computing resources, we used up to 1.5M records in our OAGS experiments. We also picked $n = 3$ for the scheme of Section~\ref{sec:evalscheme} and created three splits of 500K, 1M and 1.5M samples each, together with the three splits (one-third, two-thirds, and full) of CNNDM. Some statistics of the experimental data are shown in Table~\ref{table:datastats}. Vocabulary sizes used in each experiment are shown in its last column.

The higher vocabulary sizes of OAGS splits cause a significant difference in parameters between the two corresponding models of each method. As we can see (Table~\ref{table:scores}), Transformer models grows from 81M in CNNDM to 129M in OAGS. Another difference between the two sets of experiments is in the maximal number of encoding and decoding steps (words in source and target texts). For CNNDM, we used 400 and 100 respectively. For OAGS, we chose 200 and 50, since paper abstracts and titles should not be longer. 
%

\begin{table*}[!t]
\small 
\centering      
\setlength\tabcolsep{7pt}  
\begin{tabular}
{| l l | c c c c c c c |}
\hline
& \multicolumn{8}{c|}{\qquad \qquad \qquad \textbf{CNNDM}} \\ [0.12ex] 
\textbf{Authors} & \textbf{Models} & $\boldsymbol{\sigma_1}$ & $\boldsymbol{\sigma_2}$ & $\boldsymbol{\sigma_L}$ & $\boldsymbol{\theta}$ & $\boldsymbol{\epsilon_1}$ & $\boldsymbol{\epsilon_2}$ & $\boldsymbol{\epsilon_L}$ \\ [0.12ex] 
\hline
\multirow{2}{*}{\citeauthor{D15-1044}}
& \textsc{Abs12} & 7.13 & 18.27 & 4.54 & 37.41 & 0.19 & 0.45 & 0.12  \\ [0.07ex]
& \textsc{Abs23} & 7.64 & 21.92 & 5.93 & 63.18 & 0.12 & 0.35 & 0.094  \\ [0.77ex]
\multirow{2}{*}{\citeauthor{P17-1099}}
& \textsc{Pcov12} & 4.3 & 5.53 & 3.69 & 22.16 & 0.194 & 0.25 & 0.167  \\ [0.07ex]
& \textsc{Pcov23} & 4.46 & 9.33 & 7.03 & 37.4 & 0.119 & 0.249 & 0.188  \\ [0.77ex]
\multirow{2}{*}{\citeauthor{DBLP:journals/corr/abs-1812-02303}}
& \textsc{Nats12} & 3.6 & 9.13 & 3.47 & 81.89 & 0.044 & 0.112 & 0.042  \\ [0.07ex]
& \textsc{Nats23} & 4.53 & 16.6 & 9.74 & 92.35 & 0.049 & 0.18 & 0.106  \\ [0.77ex]
\multirow{2}{*}{\citeauthor{P18-2027}}
& \textsc{GlobEn12} & 3.53 & 8.26 & 3.96 & 19.23 & 0.184 & 0.429 & 0.206  \\ [0.07ex]
& \textsc{GlobEn23} & 4.54 & 10.15 & 4.50 & 23.2 & 0.196 & 0.437 & 0.194  \\ [0.77ex]
\multirow{2}{*}{\citeauthor{NIPS2017_7181}}
& \textsc{Trans12} & {\bf 13.53} & {\bf 38.87} & {\bf 14.92} & 11.61 & {\bf 1.166} & {\bf 3.34} & {\bf 1.285}  \\ [0.07ex]
& \textsc{Trans23} & 8.14 & 24.88 & 8.72 & {\bf 11.24} & 0.724 & 2.214 & 0.776  \\ [0.77ex]
\multirow{2}{*}{\hyperlink{P18-1063}{Chen et al.}} 
& \textsc{FastRl12} & 6.04 & 8.6 & 4.21 & 55.53 & 0.109 & 0.155 & 0.067  \\ [0.07ex]
& \textsc{FastRl23} & 4.33 & 16.87 & 6.09 & 147.78 & 0.029 & 0.114 & 0.041 \\ [0.77ex]
\multirow{2}{*}{\citeauthor{DBLP:journals/corr/PaulusXS17}}
& \textsc{Pgrl12} & 4.51 & 8.05 & 4.55 & 18.26 & 0.247 & 0.441 & 0.249  \\ [0.07ex]
& \textsc{Pgrl23} & 4.81 & 4.88 & 4.48 & 63.58 & 0.076 & 0.077 & 0.07  \\ [0.77ex]
\hline 
\end{tabular}
\caption{Data efficiency scores of the models on CNNDM experiments. $\boldsymbol{\sigma_X}$ is computed based on the corresponding $\boldsymbol{R_X}$ score. Similarly, $\boldsymbol{\epsilon_X}$ is computed based on $\boldsymbol{\sigma_X}$ and $\boldsymbol{\theta}$.}  
\label{table:effcnndm}
\end{table*}

\begin{table*}[!t]
\small 
\centering      
\setlength\tabcolsep{7pt}  
\begin{tabular}
{| l l | c c c c c c c |}
\hline
& \multicolumn{8}{c|}{\qquad \qquad \qquad \textbf{OAGS}} \\ [0.12ex] 
\textbf{Authors} & \textbf{Models} & $\boldsymbol{\sigma_1}$ & $\boldsymbol{\sigma_2}$ & $\boldsymbol{\sigma_L}$ & $\boldsymbol{\theta}$ & $\boldsymbol{\epsilon_1}$ & $\boldsymbol{\epsilon_2}$ & $\boldsymbol{\epsilon_L}$ \\ [0.12ex] 
\hline
\multirow{2}{*}{\citeauthor{D15-1044}}
& \textsc{Abs12} & 7.47 & 14.43 & 6.82 & 27.03 & 0.277 & 0.534 & 0.252  \\ [0.07ex]
& \textsc{Abs23} & 9.47 & 11.3 & 9.86 & 36.64 & 0.259 & 0.309 & 0.269  \\ [0.77ex]
\multirow{2}{*}{\citeauthor{P17-1099}}
& \textsc{Pcov12} & 2.27 & 2.21 & 4.65 & 39.84 & 0.057 & 0.055 & 0.117  \\ [0.07ex]
& \textsc{Pcov23} & 3.87 & 5.04 & 4.09 & 64.67 & 0.06 & 0.077 & 0.063  \\ [0.77ex]
\multirow{2}{*}{\citeauthor{NIPS2017_7181}}
& \textsc{Trans12} & 12.81 & {\bf 33.51} & {\bf 16.93} & 7.09 & 1.806 & {\bf 4.724} & {\bf 2.386}  \\ [0.07ex]
& \textsc{Trans23} & {\bf 16.92} & 22.3 & 14.41 & {\bf 6.63} & {\bf 2.552} & 3.364 & 2.173  \\ [0.77ex]
\multirow{2}{*}{\citeauthor{DBLP:journals/corr/PaulusXS17}}
& \textsc{Pgrl12} & 3.89 & 9.7 & 5.48 & 26.52 & 0.146 & 0.366 & 0.207  \\ [0.07ex]
& \textsc{Pgrl23} & 6.23 & 8.57 & 6.22 & 71.07 & 0.088 & 0.121 & 0.088   \\ [0.77ex]
\hline 
\end{tabular}
\caption{Data efficiency scores of the models on OAGS experiments. $\boldsymbol{\sigma_X}$ is computed based on the corresponding $\boldsymbol{R_X}$ score. Similarly, $\boldsymbol{\epsilon_X}$ is computed based on $\boldsymbol{\sigma_X}$ and $\boldsymbol{\theta}$.} 
\label{table:effOags}
\end{table*}

%
%
\subsection{Summarization Results}
\label{ssec:rouge}
%
ROUGE scores and training times (in seconds) on CNNDM experiments are shown in the middle part of Table~\ref{table:scores}. The most accurate models are \textsc{Pgrl} and \textsc{FastRl}. They both implement policy-based training and optimize w.r.t ROUGE scores. The worst performer is \textsc{Abs} and the other four fall somewhere in between, reaching similar scores with each other. 
%

The score differences between each third model and first one are usually small for all methods. We believe this has to do with the way ROUGE scores are computed. A graphical representation of the $\boldsymbol{R_1}$ trends for each method is depicted in Figure~\ref{fig:r1trends} (left). $\boldsymbol{R_2}$ and $\boldsymbol{R_L}$ (not shown) behave similarly.
%

Results on OAGS are listed on the right side of Table~\ref{table:scores}. We could not run some of the models on OAGS data. The extractive part of \textsc{FastRl} could not be easily adapted to perform word-level extraction of OAGS abstracts. Furthermore, \textsc{Nats} and \textsc{GlobEn} ran out of memory very frequently. From the remaining four, \textsc{Pgrl} is again the most accurate. \textsc{Trans} follows and \textsc{Abs} is the weakest. $\boldsymbol{R_1}$ score trends are shown in the Figure~\ref{fig:r1trends} (right).

Regarding training speed, on CNNDM we can see that \textsc{FastRl} is absolutely the best, with a considerable difference from the second (\textsc{Pgrl}). The slowest is \textsc{GlobEn} with training times at least 17x higher than those of \textsc{FastRl}. In fact, it took more than ten days to train \textsc{GlobEn} on the full CNNDM data. 
%

OAGS training times are lower than CNNDM ones, although OAGS data splits are 5.2 times bigger in number of training samples. This happens because OAGS source and target samples are actually much shorter. We see that \textsc{Pcov} is the fastest and \textsc{Trans} is the slowest. 
%
%
\subsection{Efficiency Results}
\label{ssec:dataeff}
Using Equations~\ref{eq:score}, \ref{eq:time} and \ref{eq:data} of Section~\ref{ssec:metrics} we computed the relative efficiency metrics for every method. 
%
The values for CNNDM experiments are shown in Table~\ref{table:effcnndm}. We see that \textsc{Trans} is clearly the most efficient, with highest $\sigma$, lowest $\theta$ and highest $\epsilon$. Its scores grow quickly (despite being relatively low) and training times grow slowly (despite being high) in both data intervals. 
\textsc{Pcov} and \textsc{GlobEn} manifest the slowest accuracy score gains (lowest $\sigma$), but \textsc{GlobEn} comes second in time efficiency. \textsc{Nats} on the other hand, is very time inefficient, with highest $\theta$ and lowest $\epsilon$. 
%
OAGS scores of Table~\ref{table:effOags} reflect a similar situation. \textsc{Trans} leads and \textsc{Pcov} is again the worst. The values of the other two models appear somewhere in between. 
%
%

It is not easy to explain the high score and time efficiency of \textsc{Trans}. \textsc{GlobEn} is also time efficient but not score efficient. Both of them are the biggest (highest number of parameters) and deepest (many layers) networks we tried. The exclusive feature of \textsc{Trans} is the lack of any recurrent structure. \textsc{GlobEn} and the other five make use of at least one RNN in a certain phase. It is still hasty to infer that recurrent networks hinder score efficiency or that more attention boosts it. 
%

An intuitive explanation could be the fact that in general, performance of deeper networks scales better with more data. It could also be that high-capacity networks are faster in interpreting large additions of training samples (thus low $\theta$). In fact, using more layers and bigger training datasets is what has driven the progress of deep learning solutions in many application areas. 
%

We plan to investigate this issue further in the future. One step could be to run more experiments on even bigger data sizes and smaller data intervals for checking at what point do accuracy scores keep growing. Transformer implementations with varying number of layers and other parameter setups can be further examined. 
%

Investigating data efficiency of similar solutions to tasks like QA (Question Answering, \citealp{DBLP:journals/corr/abs-1810-09580}) with standard datasets such as SQuAD \cite{rajpurkar-jia-liang:2018:Short} could also be valuable.
%
\section{Conclusions}
\label{sec:discussion}
%
In this paper, we defined three data efficiency metrics for a better evaluation of data-driven learning models. We also proposed a simple scheme for computing and reporting them, in addition to the basic accuracy scores. Text summarization and title generation tasks were chosen as a case study to see what insights the proposed scheme and metrics could reveal. For title generation, we also processed a dataset of about 35 million scientific titles and abstracts, released with this paper.
%

We applied seven recent ATS methods on the two tasks. According to our results, the two methods that mix RL concepts into the encoder-decoder framework are the fastest and the most accurate. A surprising result is the excellent efficiency of the popular Transformer model.      
%
As future work, we want to perform similar studies in analogous tasks like MT or QA. 
We would also like to investigate more in depth the Transformer model. 
\section*{Acknowledgments}
\label{sec:ack}
This research work was supported by the project No. CZ.02.2.69/0.0/0.0/16 027/0008495 (International Mobility of Researchers at Charles University) of the Operational Programme Research, Development and Education,
the project no. 19-26934X (NEUREM3) of the Czech Science Foundation and
ELITR (H2020-ICT-2018-2-825460) of the EU.

%
%

\bibliography{acl2019}
\bibliographystyle{acl_natbib}

\end{document}